# Spatial-temporal Fusion Convolutional Neural Network for Simulated Driving Behavior Recognition

Yaocong Hu
School of Automation, Southeast University
Nanjing 210096, China
Key Laboratory of Measurement and Control of Complex Systems of Engineering, Ministry of Education, Nanjing 210096, China
yaoconghu@foxmail.com

MingQi Lu
School of Automation, Southeast University
Nanjing 210096, China
Key Laboratory of Measurement and Control of Complex Systems of Engineering, Ministry of Education, Nanjing 210096, China
3012221702@qq.com

Xiaobo Lu
School of Automation, Southeast University
Nanjing 210096, China
Key Laboratory of Measurement and Control of Complex Systems of Engineering, Ministry of Education, Nanjing 210096, China
xblu2013@126.com

*Abstract*— Abnormal driving behaviour is one of the leading cause of terrible traffic accidents endangering human life. Therefore, study on driving behaviour surveillance has become essential to traffic security and public management. In this paper, we conduct this promising research and employ a two stream CNN framework for video-based driving behaviour recognition, in which spatial stream CNN captures appearance information from still frames, whilst temporal stream CNN captures motion information with pre-computed optical flow displacement between a few adjacent video frames. We investigate different spatial-temporal fusion strategies to combine the intra frame static clues and inter frame dynamic clues for final behaviour recognition. So as to validate the effectiveness of the designed spatial-temporal deep learning based model, we create a simulated driving behaviour dataset, containing 1237 videos with 6 different driving behavior for recognition. Experiment result shows that our proposed method obtains noticeable performance improvements compared to the existing methods.

## I. INTRODUCTION

Traffic security is an urgent problem all over the world. As reported by the Chinese Transport Ministry, more than 80% traffic crashes involve abnormal driving such as playing mobile phone, smoking and eating, talking with passengers etc [1]. Consequently, driving monitoring has long been a core technique in Advanced Diver Assistance System (ADAS) [2] and visual-based driving behaviour recognition has attracted great research interests in recent years.

The task of video-based driving behaviour recognition facilitates intelligent surveillance and can be regarded as a fine-grained video-based human behaviour recognition. Recognition of human behaviour in videos has received great attention in recent researches [3,4,5,6]. In [3], Karpathy *et al*. used stacked video frames as input and trained a multiple resolution CNN to recognize human behaviour. In [4], Simonyan *et al*. trained a two-stream convolutional neural networks to separately capture spatial clues and temporal clues, then combined two streams with score fusion.

Naturally, it is reasonable to transfer the spatial-temporal convolutional neural network which achieved remarkable performance in human behaviour recognition to our specific task. Therefore, in this paper, we follow the architecture of [4] to train spatial stream CNN with still frames and temporal stream CNN with multi-frame optical flow. Here, we investigate different spatial-temporal fusion strategies to further increase driving behaviour recognition accuracy.

In terms of dataset, a video-based simulated driving behaviour dataset is created. All 1237 videos were collected by a HD digital-camera involving 6 categories of normal driving, hands off the wheel, calling, playing mobile phone, smoking and talking with passenger, as depicted in Figure 1.

The contributions of this work can be categorized into three levels. First, we apply spatial-temporal CNN into fine-grained driving behaviour recognition task. Secondly, we investigate different information fusion strategies to improve the performance. Lastly, we validate the proposed method on a self-created driving behaviour dataset and achieves the state of the art.

## II. RELATED WORKS

A great may of approaches have been raised for automatic driving behaviour recognition over the last decades or so. On the basis of feature used in their approaches, we can classify them into two categories of handcrafted feature based approaches [7,8,9,10] and deep learning based approaches [11,12,14].

Zhao et al. did the original researches in driving behaviour recognition and have made remarkable contributions. They recognized driving behaviour of normal driving, operating the shift gear, eating and smoking, responding a cell phone on their private dataset. Zhao et al. in [7] designed an efficient handcrafted feature extraction method for driving behaviour recognition, which is composed of Homomorphic filter, skin segmentation and contourlet transform, and random forest was employed for final classification; in [8] adopted multi-resolution analysis to generate multi-wavelet features for behaviour recognition; in [9] introduced RBF kernel and SVM for final behaviour recognition; in [10] employed Pyramid Histogram Oriented

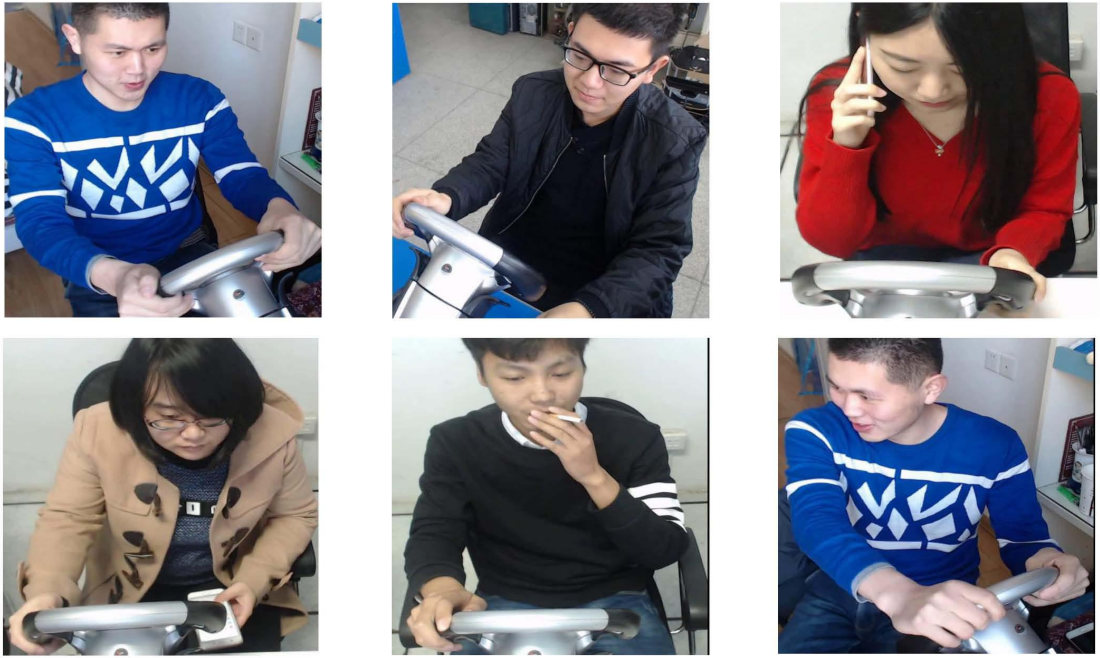

Figure 1. Examples of our self-crated dataset.

Gradient (PHOG) as spatial scale features and selected multi-layer perceptron (MLP) as classifier.

Recently, due to the breakthrough of CNN in many pattern recognition researches, deep learning based driving behaviour recognition has obtained certain popularity. Yan *et al.* in [11] first employed end-to-end deep architecture (AlexNet) for feature representation and behaviour classification. In [12], Le *et al.* adopted Faster-RCNN [13] architecture to detect cell-phone usage and driving with hands off the wheel. Koesdwiady *et al.* in [14] used deeper VGG19 architecture to learn behaviour categories with softmax supervision.

To sum up the existing driving behaviour recognition approaches, most of them are image-based and focus on appearance feature extraction from still images. In addition, standard convolutional neural network can automatically capture appearance features with end-to-end learning and achieved significant accuracy improvements compared to the handcrafted feature-based methods. However, image-based recognition ignores the inter-frame motion information which is equally essential to driving behaviour recognition.

## III. Proposed methods

This section details the core techniques of our proposed driving behavior recognition method. Firstly, we give an introduction of the proposed architecture of spatial-temporal convolutional neural network; then, we investigate and compare different spatial-temporal fusion strategy; lastly, we elaborate how to train the spatial-temporal convolutional neural network for video-based driving behavior recognition.

### A. Network Architecture

The structures of spatial stream convolutional neural network and temporal stream convolutional neural network are illustrated in Figure 2.

Spatial stream CNN can powerfully learn appearance information such as texture, contour, interest point etc from still images and it has identical structure of 16 layers VGGNet [15], which takes $224 \times 224 \times 3$ still frames as import and outputs the probability distribution of different driving behaviour categories. The spatial stream CNN is staked by thirteen convolution layers and three iner producted layers. Thirteen convolutional layers filter images with kernel size of $3 \times 3$, stride of 1 and are stacked into 5 blocks. Max pooling follows after each convolutional block with kernel size of $2 \times 2$, stride of 2. Fully connected layers can map the convolutional feature map to feature vectors. The last iner producted layer with softmax classifier has 6 neurons which output the probability distribution of different driving behaviour categories.

Temporal stream CNN can effectively learn motion information and predict the probability of different driving behaviour categories, which shares the similar structure with the spatial stream CNN except that it takes stacked optical flow with size of $224 \times 224 \times 2L$ as input. The stack of optical flow displacement between adjacent $L$ video frames can be computed by the method of Brox *et al.*'s [16]. Figure 3 show the horizontal projection $d^u$ and the vertical projection $d^v$ of the displacement vector fields between two frames. The stacked optical flow of $L$ ( $L=10$ ) consecutive frames forms a $2L$ channels input, which is calculated as follow.

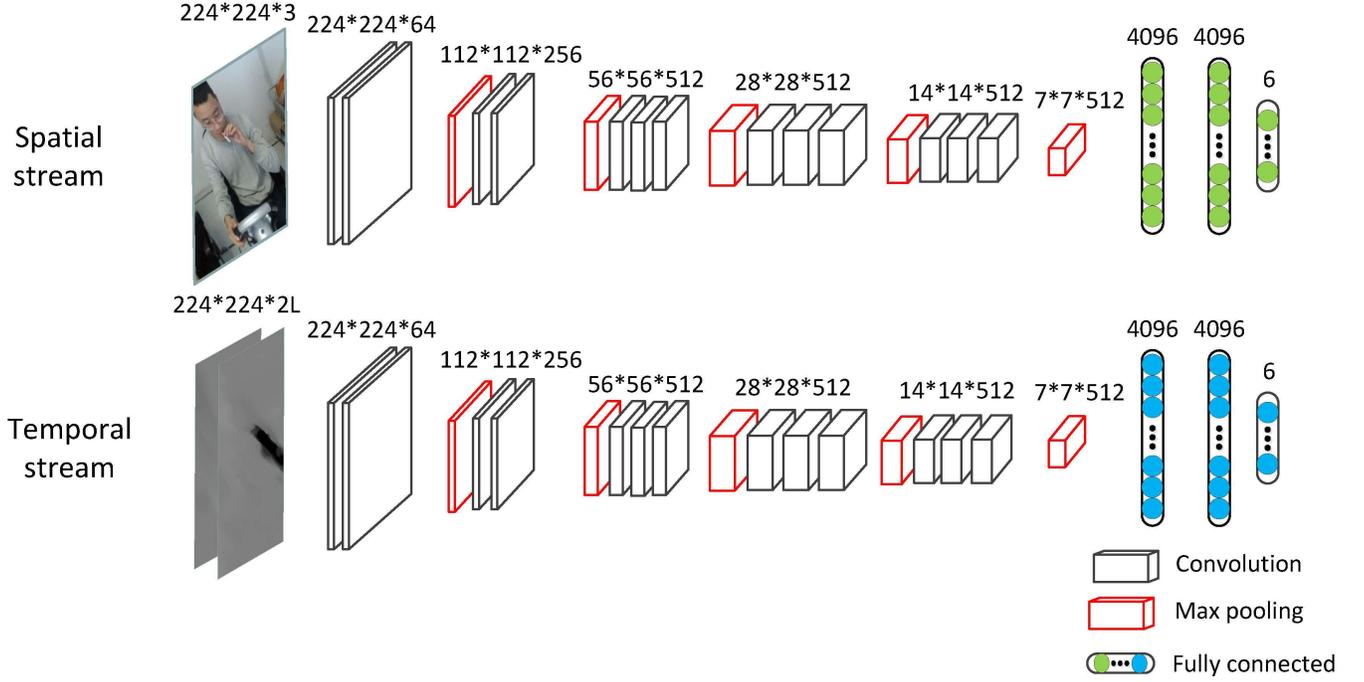

Figure 2. Spatial stream CNN and temporal stream CNN for video-based driving behaviour recognition.

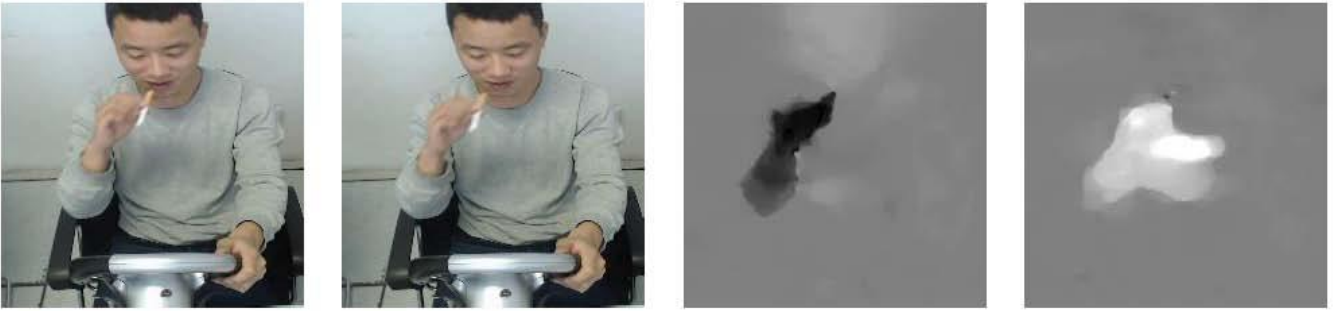

Figure 3. Examples of optical flow displacement.

$$\begin{aligned} I_T(i,j,2k-1) &= d^u_{\tau+k-1}(i,j), \\ I_T(i,j,2k) &= d^v_{\tau+k-1}(i,j), \\ i &= [1;R], j=[1,C], k=[1;L], \end{aligned} \quad (1)$$

where $I_T \in \Re^{R \times C \times 2L}$ is the input of temporal stream CNN, $R$, $C$ and $L$ represent the row, column and frame length of a video.

B. Fusion strategies

Given still frames and their pre-computed optical flow, spatial stream learns intra frame static clues, temporal stream learns inter frame dynamic clues, and each stream respectively predicts the probability of different categories. However, how to combine the spatial stream and temporal stream for final recognition needs further investigation. Here, we explore three different spatial-temporal fusion strategies to combine two streams.

In early fusion strategy, spatial stream and temporal stream are combined at the beginning of the network. still frames and their stacked optical flow are combined together giving rise to a $3+2L$ channels input (3 channels from a still frame and $2L$ channels from stacked optical flow). Then, single stream network (VGG16) takes the stacked $3+2L$ channels image as input, extract spatial-temporal features and classify different driving behaviour categories. Figure 4(a) shows the illustration of early fusion pipeline.

Contrast to early fusion strategy, still frames and their stacked optical flow are respectively fed to spatial stream and temporal stream in mid-level fusion strategy. Two streams respectively extract appearance features and motion features. Then the extracted appearance feature vector and motion feature vector are linearly combined to form a 8192-d spatial-temporal feature vector. The fused spatial-temporal feature can be denoted as:

$$f(2d-1) = f_{st}(d),$$
$$f(2d) = f_{tp}(d), \quad (2)$$

where $f_{st}$ represents the 4096-d appearance feature, $f_{tp}$ denotes the 4096-d motion feature, and $f$ is the fused spatial-temporal feature. Note that the fused spatial-temporal feature $f$ is normalized by the following equation:

$$\hat{f}(d) = \frac{f_d}{\sqrt{\sum_{d'=1}^{8192} f_{d'}^2}}, \quad (3)$$

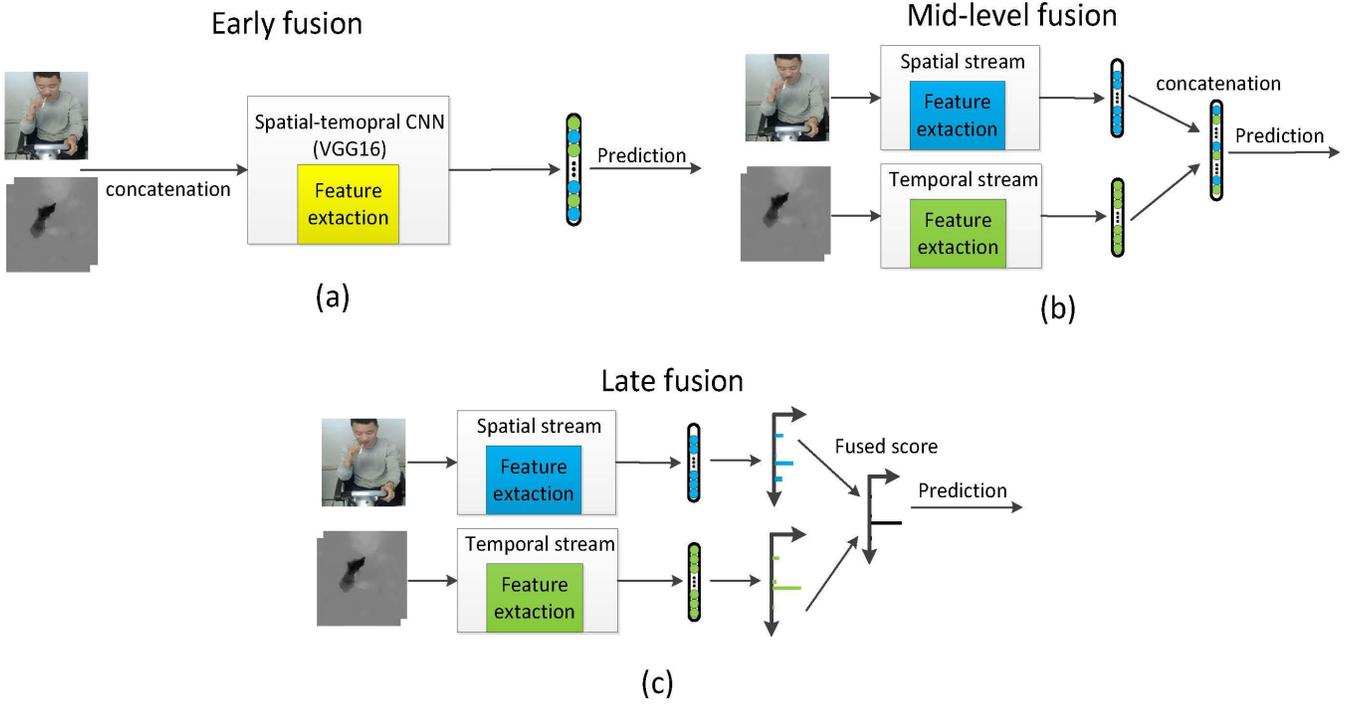

Figure 4. Illustration of different fusion strategies.

Lastly, the normalized $\hat{f}$ is utilized to train support vector machine for driving behaviour recognition. Figure 4(b) shows the illustration of mid-level fusion pipeline.

In terms of late fusion, softmax classifier in spatial stream and temporal stream are directly employed with score fusion, as shown in Figure 4(c). We define that $f_{st}$ is the 4096 dimension vector extracted by spatial stream CNN, $f_{tp}$ is the 4096 dimension vector extracted by temporal stream CNN. The probability $p(j|f_{st})$ and $p(j|f_{tp})$ can be computed by softmax classifier:

$$p(j|f_{st}) = s_{st}^j = \frac{exp(\theta_{st}^j \cdot f_{st})}{\sum_{j'=1}^{n} exp(\theta_{st}^{j'} \cdot f_{st})} \quad (4)$$

$$p(j|f_{tp}) = s_{tp}^j = \frac{exp(\theta_{tp}^j \cdot f_{tp})}{\sum_{j'=1}^{n} exp(\theta_{tp}^{j'} \cdot f_{tp})} \quad (5)$$

where $\theta_{st}^j$ and $\theta_{tp}^j$ represent the parameter of softmax classifier in spatial stream CNN and temporal stream CNN. $p(j|f_{st})$ and $p(j|f_{tp})$ are the posterior probability that $f_{st}$ and $f_{tp}$ belong to the *j-th* category judged by spatial or temporal stream.

According to bayesian principle, joint probability $p(j|f_{st}, f_{tp})$ is thus computed by late fusion. The formula derivation is same as the reference of [17], which is denoted as below:

$$score^j = p(j|f_{st}, f_{tp}) = \frac{p(f_{st}, f_{tp}|j)p(j)}{\sum_{j'=1}^{n} p(f_{st}, f_{tp}|j')p(j')}$$
$$= \frac{s_{st}^j \cdot s_{tp}^j / p(j)}{\sum_{j'=1}^{n} (s_{st}^j \cdot s_{tp}^j / p(j))} \quad (6)$$

where $s_{st}^j$ denotes the posterior probability of the *j-th* class computed by spatial stream CNN, $s_{tp}^j$ denotes the posterior probability of the *j-th* class computed by temporal stream CNN. $p(j)$ is priori probability of the *j-th* driving behaviour category. The fused classification score is denoted by $score^j$.

*C. Network training*

Before training the architecture, we pre-compute stacked optical flow in advance by employing the solution of [16] with OPENCV toolkit. We separately pre-train two stream on UCF101 [11], and then finetune two streams on our self-created driving behaviour dataset. However, in terms of early fusion strategy, still frames and their stacked optical flow are early combined at the beginning of the network; so, we pre-train and finetune the single stream network (VGG16) for final driving behaviour recognition.

## IV. EXPERIMENTAL RESULTS

Corresponding solutions are tested on Intel Core-I7 CPU, NVIDIA GeForce GTX 1080, and Ubuntu 16.04 operating system. The whole architecture is implemented by employing open source toolbox Caffe[18]. Both spatial stream and temporal stream are trained by employing the Stochastic Gradient Descent (SGD). Here, we detail the experiment dataset in section 3.1 and report the experimental comparisons with corresponding methods in section 3.2.

*A. Experiment dataset*

We create a video-based driving behaviour dataset containing 1237 videos, with average 15 seconds per video. All videos are captured by a logitech C920 HD camera and involve 6 simulated driving behaviour.

C0: Normal driving,

C1: Hands off the wheel,

C2: Calling,

C3: Playing mobile phone,

C4: Smoking,

C5: Talking with passengers.

For experiment, these videos are divided into training set (715 videos) and testing set(522 videos). The number of videos of different categories is listed in Table 1.

TABLE I. THE NUMBER OF VIDEOS OF DIFFERENT CATEGORIES.

| Category | Training set | Testing set | Total |
|---|---|---|---|
| Normal driving | 109 | 76 | 185 |
| Hands off the wheel | 95 | 70 | 165 |
| Calling | 142 | 105 | 247 |
| Playing mobile phone | 152 | 109 | 261 |
| Smoking | 103 | 75 | 178 |
| Talking with passengers | 114 | 87 | 201 |

*B. Comparisons with corresponding solutions*

We compare our proposed architecture with several corresponding methods, including the method of multi-resolution CNN [3] and the method of two-stream CNN [4]. Note that, we compare the performance of each driving behaviour category of different methods and list the quantitative experiment results in Table 2.

Karpathy *et al*. in [3] took stacked video frames as input and trained a multi-resolution CNN architecture to recognize human behaviour. Here, we repeat their implementation and achieve the total accuracy rate of 83.9%. Simonyan *et al.* in [4] proposed a baseline method, in which they trained a two stream AlexNet, and then combined two stream with score fusion. We test this method on our simulated driving behaviour dataset and achieve the total accuracy of 84.8%.

In this paper, we employ deeper convolutional neural network VGG16 to capture intra static features and inter dynamic features for driving behaviour recognition. We follow the method of [4] to separately train spatial stream convolutional neural network and temporal stream convolutional neural network. In test stage, spatial stream CNN obtains the total accuracy of 83.4% and temporal stream CNN achieves the total accuracy of 79.0%. Furthermore, we compare the performance of different spatial-temporal fusion strategies and our proposed spatial-temporal CNN with mid-level fusion strategy obtains the highest accuracy of 87.4% on self-created driving behaviour dataset. We futher show the confusion matrix of the proposed spatial-temporal CNN (mid-level fusion strategy) in Figure 5.

## V. CONCLUSION

In this paper, we propose a spatial-temporal CNN based method to recognize different driving behaviour in videos. Spatial stream CNN is employed to learn intra-frame appearance information and temporal stream CNN is utilized to capture inter-frame motion information; then we investigate different fusion strategy to combine spatial stream and temporal stream for final recognition. We perform experiment on self-created simulated driving behaviour dataset and our proposed architecture with mid-level fusion strategy achieves the state of the art recognition performance. In the future, we will further improve the performance by employing hybrid networks (CNN+ LSTM) and speed up the driving behaviour recognition to real-time. In addition, creating realdriving behaviour dataset is essential for practical applications.

## ACKNOWLEDGMENTS

This work was supported by the National Natural Science Foundation of China (No.61871123), Key Research and Development Program in Jiangsu Province (No.BE2016739) and a Project Funded by the Priority Academic Program Development of Jiangsu Higher Education Institutions.

TABLE II. ACCURACY OF THE PROPOSED SOLUTIONS AND ITS QUANTITATIVE CONTRAST WITH CORRESPONDING SOLUTIONS.

| Method | C0 | C1 | C2 | C3 | C4 | C5 | Total |
|---|---|---|---|---|---|---|---|
| Multi-resolution CNN [3] | 85.9% | 82.8% | 84.6% | 83.4% | 81.5% | 84.1% | 83.9% |
| Two stream CNN (AlexNet) [4] | 87.8% | 85.2% | 84.9% | 85.1% | 81.2% | 85.3% | 84.8% |
| Spatial stream CNN (VGG16) | 86.3% | 81.0% | 85.5% | 82.6% | 80.8% | 84.6% | 83.4% |
| Temporal stream CNN (VGG16) | 78.8% | 76.5% | 78.4% | 80.2% | 78.6% | 81.5% | 79.0% |
| Spatial-temporal CNN (early fusion) | 86.6% | 81.5% | 85.2% | 82.3% | 81.6% | 85.2% | 83.7% |
| Spatial-temporal CNN (mid-level fusion) | 89.4% | 87.9% | 87.2% | 87.9% | 85.6% | 86.5% | **87.4%** |
| Spatial-temporal CNN (late fusion) | 88.7% | 87.1% | 86.5% | 88.1% | 84.9% | 87.2% | 86.9% |

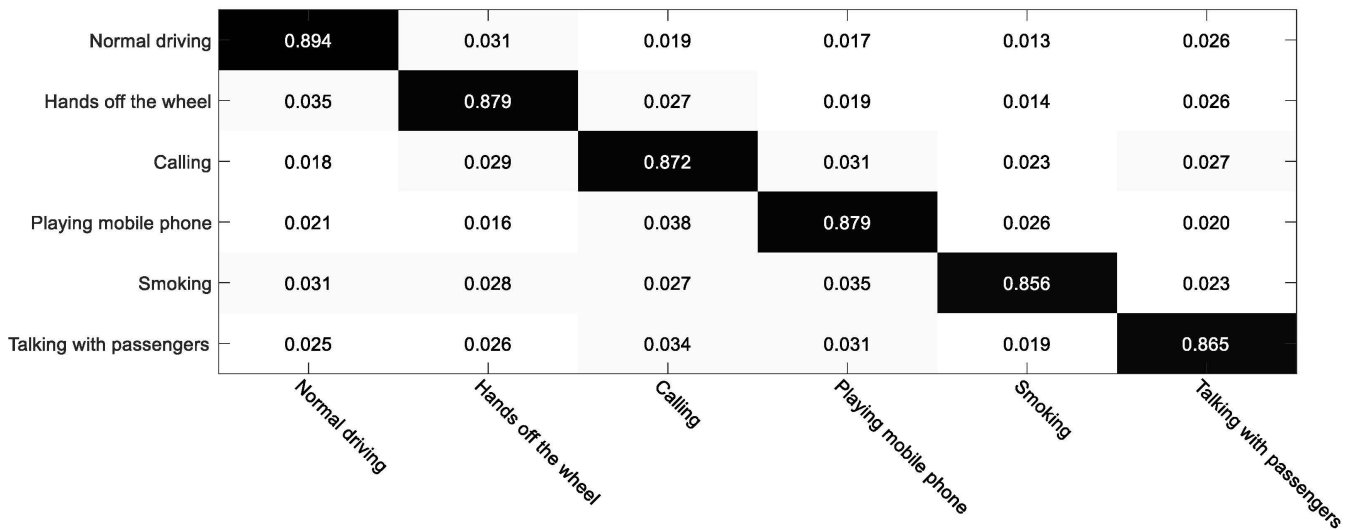

Figure 4. Confusion matrix of spatial-temporal CNN with mid-level fusion strategy.

REFERENCES

[1] Y. Yanbin, Z. Lijuan, L. Mengjun and S. Ling, "Early Warning of Traffic Accident in Shanghai Based on Large Data Set Mining," 2016 International Conference on Intelligent Transportation, Big Data & Smart City (ICITBS), Changsha, 2016, pp. 18-21. DOI= 10.1109/ICITBS.2016.149.

[2] Ba, Y., Zhang, W., Wang, Q., Zhou, R., Ren, C.:Crash prediction with behavioral and physiological features for advanced vehicle collision avoidance system. Transportation Research Part C: Emerging Technologies 74, 22-33 (2017). DOI= https://doi.org/10.1016/j.trc.2016.11.009.

[3] A. Karpathy, G. Toderici, S. Shetty, T. Leung, R. Sukthankar and L. Fei-Fei, "Large-Scale Video Classification with Convolutional Neural Networks," 2014 IEEE Conference on Computer Vision and Pattern Recognition, Columbus, OH, 2014, pp. 1725-1732. DOI= 10.1109/CVPR.2014.223.

[4] Simonyan, K., Zisserman, A.: Two-stream convolutional networks for action recognition in videos. In: Proceedings of the 27th International Conference on Neural Information Processing Systems - Volume 1, NIPS'14, pp. 568-576. MIT Press, Cambridge, MA, USA (2014). URL= http://dl.acm.org/citation.cfm?id=2968826.2968890.

[5] L. Sun, K. Jia, D. Y. Yeung and B. E. Shi, "Human Action Recognition Using Factorized Spatio-Temporal Convolutional Networks," 2015 IEEE International Conference on Computer Vision (ICCV), Santiago, 2015, pp. 4597-4605. DOI= 10.1109/ICCV.2015.522.

[6] S. Zhao, Y. Liu, Y. Han, R. Hong, Q. Hu and Q. Tian, "Pooling the Convolutional Layers in Deep ConvNets for Video Action Recognition," in IEEE Transactions on Circuits and Systems for Video Technology. DOI= 10.1109/TCSVT.2017.2682196.

[7] C. H. Zhao, B. L. Zhang, J. He and J. Lian, "Recognition of driving postures by contourlet transform and random forests," in IET Intelligent Transport Systems, vol. 6, no. 2, pp. 161-168, June 2012.DOI= 10.1049/iet-its.2011.0116.

[8] Chihang Zhao, Yongsheng Gao, Jie He, Jie Lian, Recognition of driving postures by multiwavelet transform and multilayer perceptron classifier, Engineering Applications of Artificial Intelligence, Volume 25, Issue 8, 2012, Pages 1677-1686, DOI= https://doi.org/10.1016/j.engappai.2012.09.018.

[9] C. Zhao, B. Zhang, J. Lian, J. He, T. Lin and X. Zhang, "Classification of Driving Postures by Support Vector Machines,"

2011 Sixth International Conference on Image and Graphics, Hefei, Anhui, 2011, pp. 926-930. DOI= 10.1109/ICIG.2011.184.

[10] Zhao, C.H., Zhang, B.L., Zhang, X.Z., Zhao, S.Q., Li, H.X, Erratum to: Recognition of driving postures by combined features and random subspace ensemble of multilayer perceptron classifiers. Neural Computing and Applications 22(1), 175-184 (2013). DOI= 10.1007/s00521-012-1121-0.

[11] Chao Yan, B. Zhang and F. Coenen, "Driving posture recognition by convolutional neural networks," 2015 11th International Conference on Natural Computation (ICNC), Zhangjiajie, 2015, pp. 680-685. DOI= 10.1109/ICNC.2015.7378072.

[12] Koesdwiady, A., Bedawi, S.M., Ou, C., Karray, F.: End-to-end deep learning for driver distraction recognition. In: F. Karray, A. Campilho, F. Cheriet (eds.) Image Analysis and Recognition, pp. 11-18. Springer International Publishing, Cham (2017)

[13] Ren, S., He, K., Girshick, R., Sun, J.: Faster r-cnn: Towards real-time object detection with region proposal networks. In: Proceedings of the 28th International Conference on Neural Information Processing Systems - Volume 1, NIPS'15, pp. 91-99. MIT Press, Cambridge, MA, USA (2015). URL= http://dl.acm.org/citation.cfm?id=2969239.2969250.

[14] T. H. N. Le, Y. Zheng, C. Zhu, K. Luu and M. Savvides, "Multiple Scale Faster-RCNN Approach to Driver's Cell-Phone Usage and Hands on Steering Wheel Detection," 2016 IEEE Conference on Computer Vision and Pattern Recognition Workshops (CVPRW), Las Vegas, NV, 2016, pp. 46-53. doi: 10.1109/CVPRW.2016.13.

[15] K. Simonyan and A. Zisserman. Very Deep Convolutional Networks for Large-Scale Image Recognition. In ICLR, 2015.

[16] Brox, T., Bruhn, A., Papenberg, N., Weickert, J.: High accuracy optical flow estimation based on a theory for warping. Computer Vision - ECCV 2004: 8th European Conference on Computer Vision, Prague, Czech Republic, May 11-14, 2004. Proceedings, Part IV, pp. 25-36 (2004)

[17] Pengjie Tang, Hanli Wang, Sam Kwong, G-MS2F: GoogLeNet based multi-stage feature fusion of deep CNN for scene recognition, Neurocomputing, Volume 225, 2017, 188-197, DOI= https://doi.org/10.1016/j.neucom.2016.11.023.

[18] Jia, Y., Shelhamer, E., Donahue, J., Karayev, S., Long, J., Girshick, R., Guadarrama, S., Darrell, T.: Caffe: Convolutional architecture for fast feature embedding. MM 2014 - Proceedings of the 2014 ACM Conference on Multimedia (2014).